\documentclass[sigconf,10pt]{acmart}

\usepackage{amsmath,amssymb,amsfonts}
\usepackage{graphicx}
\usepackage{textcomp}
\usepackage{xcolor}
\usepackage{hyperref}

\usepackage[normalem]{ulem}

\usepackage{float}
\usepackage{subfig}
\usepackage{wrapfig}
\usepackage{lipsum} 
\usepackage{textcomp}
\usepackage{tikz}
\usepackage{comment}
\usepackage{colortbl}   
\usepackage{soul}
\soulregister{\cite}7
\soulregister{\citep}7 
\soulregister{\citet}7 
\soulregister{\ref}7
\soulregister{\pageref}7
\soulregister{\brown}7
\soulregister{\red}7

\usepackage[linesnumbered,ruled,vlined]{algorithm2e}
\usepackage{algpseudocode}
\usepackage{amsmath}

\usepackage{booktabs}  
\usepackage{multirow}  
\usepackage{threeparttable}  
\usepackage{makecell}        
\usepackage{pifont}     
\usepackage{adjustbox}
\usepackage{ragged2e}

\usepackage{enumitem}

\newenvironment{myquote}[1]%
  {\list{}{\leftmargin=#1\rightmargin=#1}\item[]}%
  {\endlist}

\setcopyright{none}
\renewcommand\footnotetextcopyrightpermission[1]{}

\begin{document}

\newcommand{\GY}[1]{\textcolor{blue}{Geng: #1}}
\newcommand{\todo}[1]{\textcolor{red}{\sf\bfseries Todo: #1}}
\newcommand{\bred}[1]{\textcolor{red}{\sf\bfseries #1}}
\newcommand{\red}[1]{\textcolor{red}{#1}}
\newcommand{\blue}[1]{\textcolor{blue}{#1}}
\newcommand{\yellow}[1]{\textcolor{yellow}{#1}}
\newcommand{\purple}[1]{\textcolor{purple}{#1}}
\newcommand{\brown}[1]{\textcolor{brown}{#1}}
\newcommand{\cross}[1]{\textcolor{red}{\sout{#1}}}
\newcommand{\SL}[1]{\textcolor{purple}{Sheng: #1}}
\newcommand{\YD}[1]{\textcolor{teal}{Yue: #1}}

\newcommand{\xulong}[1]{\COMMENT{\textcolor{cyan}{\sf\bfseries Xulong: #1}}}

\newcommand{\squishlist}{
    \begin{list}{$\bullet$}
        { \setlength{\itemsep}{0pt}      \setlength{\parsep}{0pt}
            \setlength{\topsep}{0.5pt}       \setlength{\partopsep}{0pt}
            \setlength{\listparindent}{-2pt}
            \setlength{\itemindent}{-5pt}
            \setlength{\leftmargin}{0.75em} \setlength{\labelwidth}{0em}
            \setlength{\labelsep}{0.2em} } }
    
\newcommand{\squishend}{
\end{list}  }

\newcommand{\yw}[1]{\textcolor{blue}{[yw:~#1]}}


\title{EdgeOL: Efficient in-situ Online Learning on Edge Devices}

\author{Sheng Li}
\affiliation{%
  \institution{University of Pittsburgh}
  \city{Pittsburgh}
  \state{PA}
  \country{USA}}
\email{shl188@pitt.edu}

\author{Geng Yuan}
\affiliation{%
  \institution{University of Georgia}
  \city{Athens}
  \state{GA}
  \country{USA}}
\email{geng.yuan@uga.edu}

\author{Yue Dai}
\affiliation{%
  \institution{University of Pittsburgh}
  \city{Pittsburgh}
  \state{PA}
  \country{USA}}
\email{yud42@pitt.edu}

\author{Tianyu Wang}
\affiliation{%
  \institution{University of Pittsburgh}
  \city{Pittsburgh}
  \state{PA}
  \country{USA}}
\email{tiw81@pitt.edu}

\author{Yawen Wu}
\affiliation{%
  \institution{University of Pittsburgh}
  \city{Pittsburgh}
  \state{PA}
  \country{USA}}
\email{yawen.wu@pitt.edu}


\author{Alex K. Jones}
\affiliation{%
  \institution{University of Pittsburgh}
  \city{Pittsburgh}
  \state{PA}
  \country{USA}}
\email{akjones@pitt.edu}

\author{Jingtong Hu}
\affiliation{%
  \institution{University of Pittsburgh}
  \city{Pittsburgh}
  \state{PA}
  \country{USA}}
\email{jthu@pitt.edu}

\author{Tony (Tong) Geng}
\affiliation{%
  \institution{University of Rochester}
  \city{Rochester}
  \state{NY}
  \country{USA}}
\email{tong.geng@rochester.edu}

\author{Yanzhi Wang}
\affiliation{%
  \institution{Northeastern University}
  \city{Boston}
  \state{MA}
  \country{USA}}
\email{yanzhiwang@northeastern.edu}

\author{Bo Yuan}
\affiliation{%
  \institution{Rutgers University}
  \city{New Brunswick}
  \state{NJ}
  \country{USA}}
\email{bo.yuan@soe.rutgers.edu}

\author{Yufei Ding}
\affiliation{%
  \institution{UCSD}
  \city{San Diego}
  \state{CA}
  \country{USA}}
\email{yufeiding@ucsd.edu}

\author{Xulong Tang}
\affiliation{%
  \institution{University of Pittsburgh}
  \city{Pittsburgh}
  \state{PA}
  \country{USA}}
\email{xulongtang@pitt.edu}



\begin{abstract}
Emerging applications, such as robot-assisted eldercare and object recognition, 
generally employ deep learning neural networks (DNNs) and naturally require: i) handling streaming-in inference requests and ii) adapting to possible deployment scenario changes.
Online model fine-tuning is widely adopted to satisfy these needs. However, an inappropriate fine-tuning scheme could involve significant energy consumption, making it challenging to deploy on edge devices.
In this paper, we propose EdgeOL, an edge online learning framework that optimizes inference accuracy, fine-tuning execution time, and energy efficiency 
through both inter-tuning and intra-tuning optimizations. 
Experimental results show that, on average, EdgeOL reduces overall fine-tuning execution time by 64\%, energy consumption by 52\%, and improves average inference accuracy by 1.75\% over the immediate online learning strategy.
\end{abstract}




\maketitle 
\pagestyle{plain} 


\section{Introduction}\label{sec:intro}
\label{sec:intro}


With the exceptional performance, Deep learning neural networks (DNNs) have gained significant popularity in emerging application domains such as robot-assisted eldercare~\cite{do2018rish, bemelmans2012socially}, object recognition~\cite{doshi2020continual, pellegrini2020latent}, and wild surveillance~\cite{akbari2021applications, ke2020smart}. 
These cutting-edge applications generally deploy DNN models on energy-constrained edge devices, such as robots and internet-of-things (IoT) devices~\cite{lesort2020continual, zhao2021general, chen2019deep, li2018learning, wang2018deep}.

There are two fundamental requirements of deploying DNN models on edge devices: (1)~{\it adaptiveness} and (2)~{\it energy-efficiency}.
From the perspective of {\it adaptiveness}, 
the DNN applications commonly have streaming-in training data and inference requests over time. 
This requires model {\it fine-tuning} using incoming training data to i) adapt to scenario changes, 
and meanwhile ii) maximize the average inference accuracy for the streaming-in inference requests (details in Section \ref{sec:background}). 
For example, to maintain high accuracy, an object recognition system needs a timely update of its DNN model when working under different environments or conditions, such as involving new classes of data or instances of existing classes but with new patterns
(e.g., different illumination conditions, background, and occlusion)~\cite{lomonaco2017core50,lomonaco2020rehearsal, she2020openloris, roy2023cl3, hao2023enhanced} while keeping the recognition functions online. 
Regarding {\it energy efficiency}, the DNN applications need to optimize energy efficiency since they are often deployed on edge devices with constrained power capacities~\cite{wang2019e2, yang2017designing, wu2021enabling}, such as battery-powered robots, mobile phones, or IoT devices.
Existing approaches usually employ online learning to ensure model adaptiveness.
Immediate online learning is an extreme case of online learning that performs immediate model fine-tuning once new training data arrives~\cite{nallaperuma2019online, gomes2019machine, fontenla2013online, hayes2020lifelong}. As a result, it guarantees high inference accuracy for incoming inference requests since the model is always up-to-date. 
However, this requires a large amount of computation as well as significant overheads from frequent model loading, saving, and system initialization like model compilation, making it less energy efficient.
On the other hand, fine-tuning (i.e., training) models at a fixed and lower frequency seems to be a reasonable trade-off for accuracy and energy efficiency. But it is still far from an optimal solution (details are discussed in Section \ref{sec:moti_inter}).

In this paper, we propose {\tt EdgeOL}, an online learning framework for edge devices, aiming to achieve both adaptiveness and energy efficiency. Our design motivation stems from the observation that there are redundant computation and memory accesses during the model fine-tuning stage in existing online learning approaches. Specifically, we first observe that many fine-tuning rounds contribute little to the inference accuracy. That is, selectively delaying and merging some fine-tuning rounds and reducing the fine-tuning frequency will not hurt the inference accuracy. We call this {\it inter-tuning redundancy}. Second, we observe that some layers gradually reach convergence during fine-tuning. In this case, freezing those converged layers will not affect the inference accuracy. We call this {\it intra-tuning redundancy}. 
Moreover, freezing layers helps to avoid over-adaptation to the training data and improve the model convergence speed by reducing the number of weights being trained.
This allows the streaming-in inference requests to use a robust model with higher accuracy while reducing the fine-tuning time and energy consumption.

To summarize, we make the following contributions.

\squishlist
    \item We conduct a comprehensive characterization that quantifies the fine-tuning execution time, energy consumption, and inference accuracy of existing online learning approa-
    ches. We reveal that there exist substantial inter-tuning and intra-tuning redundancies that can be optimized to significantly reduce the fine-tuning execution time and energy consumption while improving the inference accuracy.
    \item We propose {\tt EdgeOL} framework that consists of: i) inter-tuning optimization that dynamically and adaptively determines the fine-tuning frequency, and ii) similarity-guided layer freezing for intra-tuning optimization.
    \item We evaluate {\tt EdgeOL} using various DNN models and datasets in both computer vision (CV) and natural language processing (NLP) domains. 
    Experimental results show that, compared to immediate online learning in CV domain, {\tt EdgeOL} saves 64\% (67\% in NLP domain) of overall fine-tuning execution time and 52\% (54\% in NLP domain) of energy consumption on average. 
    Furthermore, {\tt EdgeOL} improves the average inference accuracy of all streaming-in inference requests by 1.75\% (1.52\% in NLP domain).
    \item We demonstrate {\tt EdgeOL} outperforms state-of-the-art efficient training methods, including layer freezing frameworks i) Egeria and ii) SlimFit, iii) sparse training framework RigL, and iv) efficient online learning framework Ekya, even if they have been optimized by our proposed inter-tuning optimization. {\tt EdgeOL} provides 2.1$\times$, 2.2$\times$, 2.8$\times$, and 2.0$\times$ energy savings, respectively, while delivering 1.78\%, 2.18\%, 2.33\%, and 1.50\% higher accuracy.
\squishend

\section{Background}
\label{sec:background}

{\bf Scenario change.}
The deployment scenario of an already-in-use model may change over time as the user usage scenario evolves~\cite{logacjov2021learning, liu2021lifelong, irfan2021lifelong, pinto2016curious, ma2023cost}.
These changes can generally be classified into two types, the introduction of i) instances of existing data classes but with new feature patterns~\cite{hendrycks2019benchmarking, she2020openloris, lomonaco2020rehearsal} and ii) new classes of data~\cite{lomonaco2017core50, roy2023cl3, hao2023enhanced}.
Instances with new feature patterns refer to scenarios where the model encounters variations (e.g., different illumination conditions, background, and occlusion) in previously recognized data classes. These variations could be due to changes in environmental conditions, user behavior, or other factors that alter the appearance or characteristics of the data. 
On the other hand, the introduction of new classes of data presents a different challenge, where the model must learn to identify classes that were completely absent previously.
In our work, we comprehensively evaluate our method for both types of scenario changes.

\begin{figure}[t]
\centering
    \includegraphics[width=1\columnwidth]{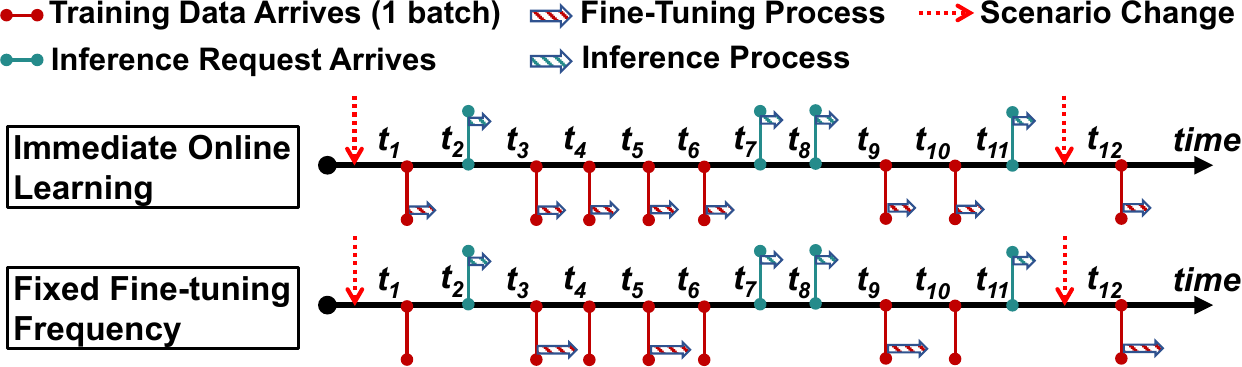}
\caption{Examples of immediate online learning and fixed fine-tuning frequency online learning.
}
\label{fig:immed_20_offl}
\end{figure}

{\bf Online learning.}
\label{sec:bg_on_off}
Conducting online learning can effectively mitigate the effect of scenario change and improve accuracy, which is essential for models to function effectively in the ever-changing real-world environment\cite{shmelkov2017incremental, bhardwaj2022ekya, rebuffi2017icarl}.
In online learning, the fine-tuning data is not well-prepared all at once but rather constantly streaming in, with new data arriving continuously throughout each scenario~\cite{wu2021enabling, pinto2016curious, lomonaco2017core50}.
Figure \ref{fig:immed_20_offl} illustrates the \textit{immediate online learning} and an example of \textit{fixed-frequency online learning} that initiates fine-tuning process after receiving a certain fine-tuning data batches (e.g., two batches in the example in the figure).
We assume two scenario changes in the figure, indicated by the red dotted arrows right before $t_1$ and $t_{12}$.
It involves eight received fine-tuning data batches represented by eight red lines, respectively.
The green lines indicate four inference requests. 
Note that, in practice, inference requests might arrive in bursts (e.g., at $t_7$ and $t_8$).
In immediate online learning, model fine-tuning is triggered once fine-tuning data (i.e., training data) arrives. Thus, the model is fine-tuned eight times in this example.
On the other hand, fixed-frequency online learning 
fine-tunes the model four times in this case. 
In general, immediate online learning achieves the highest average inference accuracy by frequently updating the model. 
However, this involves significant overheads from frequent model loading, saving, and system initialization (e.g., model compilation), making it less energy efficient.
While fixed-frequency fine-tuning seems to be a reasonable tradeoff between accuracy and energy efficiency, it lacks flexibility and adaptiveness to different cases and is far from an optimal solution (details discussed in Section \ref{sec:moti_inter}).


It is worth mentioning that, for edge online learning systems that employ a supervised learning paradigm, there are several different methods to address the data labeling issue for the newly arrived training data.
For example, some systems label the training data using a highly accurate but expensive model (with deeper architecture and a larger size)~\cite{bhardwaj2022ekya, khani2023recl, khani2021real, Mullapudi_2019_ICCV}, and this is essentially that of supervising a low-cost ``student” model with a high-cost ``teacher” model (knowledge distillation)~\cite{gou2021knowledge, cho2019efficacy, shmelkov2017incremental}. The reason why we need to train a small model is that the large model cannot keep up with inference on the edge. Moreover, the training data can also be labeled by open-source labeling platforms~\cite{soghoyan2021toolbox, chew2019smart, fiedler2019imagetagger} or stand-alone labeling service providers~\cite{humansignal2023, telusinternational2023, imerit2023}.

\begin{figure}[t]
\centering
    \includegraphics[width=1\columnwidth]{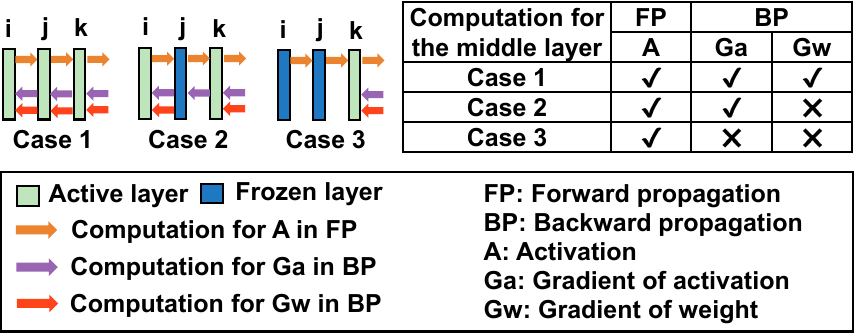}
    \caption{Computation of DNN training.}
\label{fig:computation_freezing}
\vspace{-0.1in}
\end{figure}

{\bf Average inference accuracy.}
In online learning, the ongoing model fine-tuning and continuous arrival of inference requests necessitate an evaluation metric to assess the effectiveness of fine-tuning during the entire online learning process. Thus, the average inference accuracy, which is the arithmetic mean of (instantaneous) inference accuracies for all requests, is commonly used to serve the purpose~\cite{bhardwaj2022ekya}. For example, inference requests occur at times $t_2$, $t_7$, $t_8$, and $t_{11}$, with corresponding accuracies $A_{t2}$, $A_{t7}$, $A_{t8}$, and $A_{t_{11}}$. The average accuracy is thus calculated as $(A_{t2}+A_{t7}+A_{t8}+A_{t{11}})/4$. 

{\bf Reducing computation by layer freezing.}
\label{sec:bg_training_freezing}
As shown in Figure {\ref{fig:computation_freezing}}, the computation cost in a DNN training iteration is mainly contributed by computing the activations in forward propagation and computing the gradients of weights and activations in backward propagation.
If a layer (e.g., layer$_j$) is frozen, its weights will not be updated. Thus, there is no need to calculate the weight gradients for layer$_j$ (Case 2 in Figure {\ref{fig:computation_freezing}}). 
Furthermore, if all the trainable layers before layer$_j$ ($\forall$ layer$_i | i<j$) are also frozen, then the back-propagation stops at layer$_j$; thus, there is no need to compute the activation gradient for those layers (Case 3 in Figure {\ref{fig:computation_freezing}}).

\section{Challenges and Opportunities}
\label{sec:motivation}


First, we quantitatively characterize the impact of different online learning strategies on the fine-tuning execution time, energy consumption, and inference accuracy. The differences between these learning strategies primarily target two aspects: \textbf{1) inter-tuning} and \textbf{2) intra-tuning}. 

In this section, we employ two popular DNN models Res-
Net50~\cite{he2016deep} and MobileNetV2~\cite{sandler2018mobilenetv2}, and use the NC (New Class) benchmark in the widely-used CORe50 dataset~\cite{lomonaco2017core50} as an example for testing. There are 9 scenarios in total in this benchmark and the scenarios appear one after one, each of which introduces new classes of data on top of the existing classes. The model is originally well-trained using the training data in the first scenario and will be online fine-tuned with corresponding training data and serve inference requests in each subsequent scenario (i.e., scenario 2$\sim$9). Both the training data 
and inference requests arrive continuously over time. Please refer to Section \ref{sec:motivation_inter_setup} for details of the experimental setup.

\begin{table}[t]
\setlength\tabcolsep{1.6pt}
\centering
  \scriptsize
  \caption{The configurations of different online learning strategies.}
  \begin{tabular}{m{2.8cm}<{\centering}m{0.8cm}<{\centering}m{0.66cm}<{\centering}m{0.66cm}<{\centering}m{0.66cm}<{\centering}m{0.66cm}<{\centering}m{0.66cm}<{\centering}m{0.6cm}<{\centering}}
    \toprule
    Strategy & Immed. & S1 & S2& S3& S4 & S5& S6 \\
    \midrule
    Number of batches to trigger a fine-tuning round & 1 & 5 & 10 & 20 & 50 & 100 & 200 \\
    \hline
    Number of fine-tuning rounds triggered & 6,000 & 1,200 & 600 & 300 & 120 & 60 & 30 \\
  \bottomrule
\end{tabular}
\label{strategy_table}
\end{table}


\subsection{Inter-tuning}
\label{sec:moti_inter}
Recall that the online learning approach tunes a model whenever sufficient training data arrives (e.g., a batch of training data). However, this timely and immediate fine-tuning approach consumes time and energy.
We conduct a quantitative study by varying the fine-tuning frequency. 

We consider the \textit{fine-tuning frequency} as the number of the triggered fine-tuning rounds during a certain period, where each round is triggered after a fixed number of training data batches arrive.
Therefore, with higher fine-tuning frequency, fewer batches are required to trigger a fine-tuning round.
As shown in Table \ref{strategy_table}, we choose seven different fine-tuning frequencies, from immediate online learning ({\it Immed.}) to the less frequent fine-tuning strategies $S6$.
Note that, in this experiment, the total amount of used training data is not changed as we only delay and merge some fine-tuning rounds, where a round will involve more training data. We do not skip using any training data.

\begin{figure}[t]
    \centering
    \includegraphics[width=0.98\columnwidth]{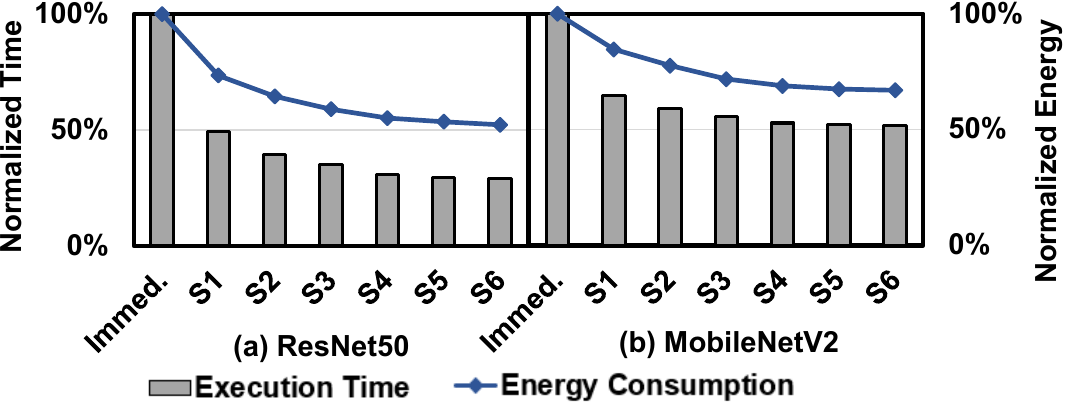}
    \caption{Overall fine-tuning execution time and energy consumption of seven strategies.}
    \label{fig:moti_inter_time_energy}
\end{figure}

\begin{figure}[t]
    \centering
    \includegraphics[width=0.95\columnwidth]{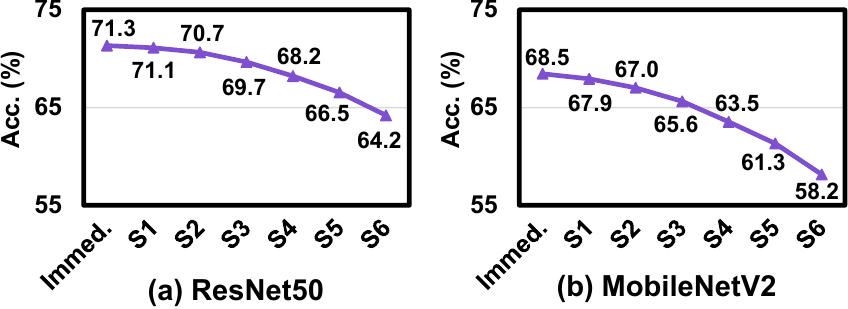}
    \caption{Average inference accuracy using seven strategies.}
    \label{fig:moti_inter_acc}
\end{figure}

\label{sec:moti_inter_observe}
\textbf{Execution time, energy, and accuracy.}
We investigate the overall fine-tuning execution time and energy consumption of the seven strategies during the entire online learning process.
The results are normalized to \textit{Immed}.
As one can observe from Figure \ref{fig:moti_inter_time_energy}, 
the fewer number of fine-tuning rounds (i.e., from \textit{Immed} to \textit{S6}), the more savings in overall execution time and energy consumption. 
This is because frequently triggering fine-tuning introduces significant overheads such as model loading, saving, and system initialization (e.g., model compilation). 
However, the savings in time and energy come with accuracy degradation, as shown in Figure~\ref{fig:moti_inter_acc}.
Therefore, an important question we seek to answer is that \textit{How can we reduce the fine-tuning frequency while maintaining the accuracy?}

\begin{figure}[t]
    \centering
    \includegraphics[width=0.95\columnwidth]{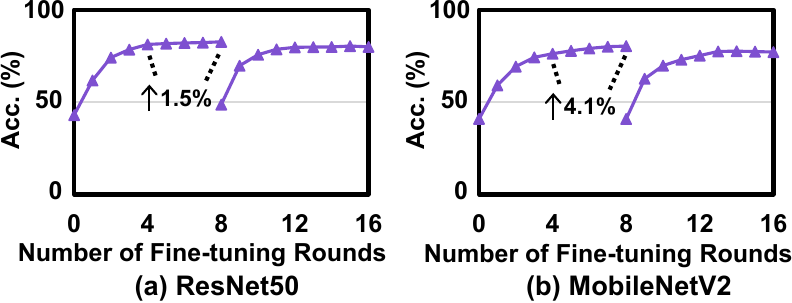}
    \caption{Accuracy improvement curve of ResNet50 and MobileNetV2 in two consecutive scenarios using strategy $S5$.}
    \label{fig:moti_inter_acc_curve}
\end{figure}

\textbf{Accuracy improvement by each fine-tuning round.}
\label{sec:accuracy-each-fine-tuning-round}
To answer the above question, 
Figure~\ref{fig:moti_inter_acc_curve} takes strategy $S5$ as an example (other strategies showed similar trends) and shows the model validation accuracy\footnote{Details of validation accuracy are defined in the Section~\ref{sec:scheme_freq_design}} over fine-tuning rounds in the two consecutive scenarios. One can make the following observations from the results. First, as expected, there is a significant accuracy drop when there is a scenario change. Second, after the scenario change, the validation accuracy improves quickly in early fine-tuning rounds and saturates in later rounds. This demonstrates that not every fine-tuning round contributes significantly to the accuracy. Therefore, one can potentially delay and merge some of the fine-tuning rounds without affecting the accuracy. Third, the number of fine-tuning rounds where accuracy saturates varies across models, as indicated by the two models in Figure~\ref{fig:moti_inter_acc_curve}, implying the need for an adaptive approach to determine the fine-tuning rounds that could be delayed and merged.

\subsection{Intra-tuning}
\label{sec:moti_intra}

For a given round of fine-tuning, we investigate if we can further reduce computation by studying different layers' impacts on accuracy.
A recent study has revealed that some DNN layers show a higher representational similarity even if the models are trained on different datasets~\cite{kornblith2019similarity}. 
Inspired by this, we explore if we can reduce computation costs without compromising accuracy by ``{\it freezing}'' some layers.
\noindent\textbf{A Preliminary Exploration.}
\label{sec:Preliminary_Exploration}
We conduct a preliminary experiment to show the effect of layer freezing on execution time, energy consumption, and accuracy.
As an example, we employ ResNet50 and MobileNetV2 on the NC benchmark and follow the experimental setup in Section~\ref{sec:motivation_inter_setup}, and we use the strategy $S5$.
\setlength{\columnsep}{8pt} 
\begin{wrapfigure}{R}{0.21\textwidth}
 \centering
\includegraphics[width=0.21\textwidth]{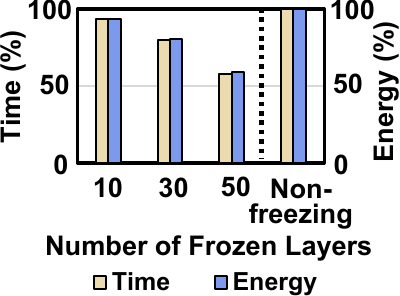}
    \caption{Fine-tuning time and energy when freezing different numbers of layers.}
\label{fig:moti_intra_time_energy}
\end{wrapfigure}
First, we randomly freeze 10, 30, and 50 CONV layers (with the corresponding BN layers) throughout the entire online learning process.

Figure~\ref{fig:moti_intra_time_energy} depicts the execution time and energy consumption when freezing different numbers of layers of ResNet50. 
We normalized the results to a non-freezing baseline.
One can observe that layer freezing effectively reduces the time and energy of the online learning process and the savings increase as more layers are frozen. 
Importantly, these time and energy savings can be directly achieved using native DNN training frameworks (e.g., PyTorch) and do \textit{not} require any support of dedicated libraries (e.g., sparse computation) or specific hardware accelerators.

\begin{figure}[htpb]
 \centering
\includegraphics[width=0.9\columnwidth]{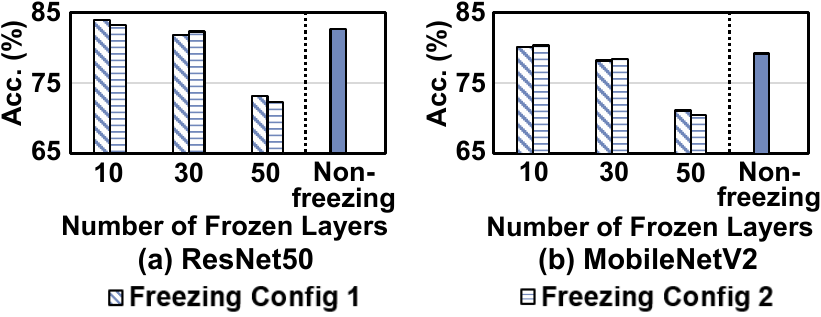}
\caption{Average inference accuracy when freezing different numbers of layers.}
\label{fig:moti_intra_acc}
\end{figure}

Figure~\ref{fig:moti_intra_acc} shows the accuracy when freezing different numbers of layers of ResNet50 and MobileNetV2.
For a more comprehensive exploration, we randomly selected two groups of layers to freeze, denoted as Config 1 and Config 2, respectively.
First, we observe that as more layers are frozen, the accuracy drops accordingly.
However, when we compare the accuracy of the two configurations, even with the same number of frozen layers, there is still a significant difference in accuracy.
The most intriguing observation is that, when freezing layers appropriately (e.g., freezing 10 layers), the accuracy is not decreased and, in fact, increases over the non-freezing baseline.
This is because freezing layers can i) avoid over-adaptation to training data and ii) improve model convergence speed (details in Figure \ref{fig:eval_acc_curve} and Section \ref{sec:eval_overall}).
This indicates that layer freezing is not a simple trade-off between accuracy and efficiency. 
As the preliminary experimental results, we use relatively simple settings to explore the feasibility and potential of incorporating layer freezing in online learning. 
Note that similar trends are observed with other experimental settings (e.g., using other online learning strategies with different fine-tuning frequencies).

Even though layer freezing has promising performance in time and energy efficiency and accuracy in online learning, several critical questions remain.
Firstly, to incorporate layer freezing in the fine-tuning process, determining which layers are appropriate to freeze is a challenging problem.
Secondly, there are also many choices about when to freeze a layer.
In addition, when the scenario changes, we also need to decide whether to resume training on certain frozen layers to quickly adapt to the new scenario.

\section{EdgeOL Design}
\label{sec:scheme}

\begin{table}[b]
\centering
  \footnotesize
 \caption{Abbreviation Description.}
  \begin{tabular}{m{1.75cm}m{6.05cm}}
    \toprule
    Abbreviation&Description\\
    \midrule
    batches\_ava             &  Number of data batches available for fine-tuning\\
    batches\_needed          &  Number of data batches needed to trigger a fine-tuning round  \\
    freeze\_interval        &  The interval (iterations) to conduct a freezing process \\
    CKA\_variation          &  The variation rate of CKA \\
    CKA\_TH                 &  CKA variation rate threshold to regard CKA is stable\\
  \bottomrule
\end{tabular}
\label{abbr}
\end{table}

\begin{figure}[t]
 \centering
\includegraphics[width=1\columnwidth]{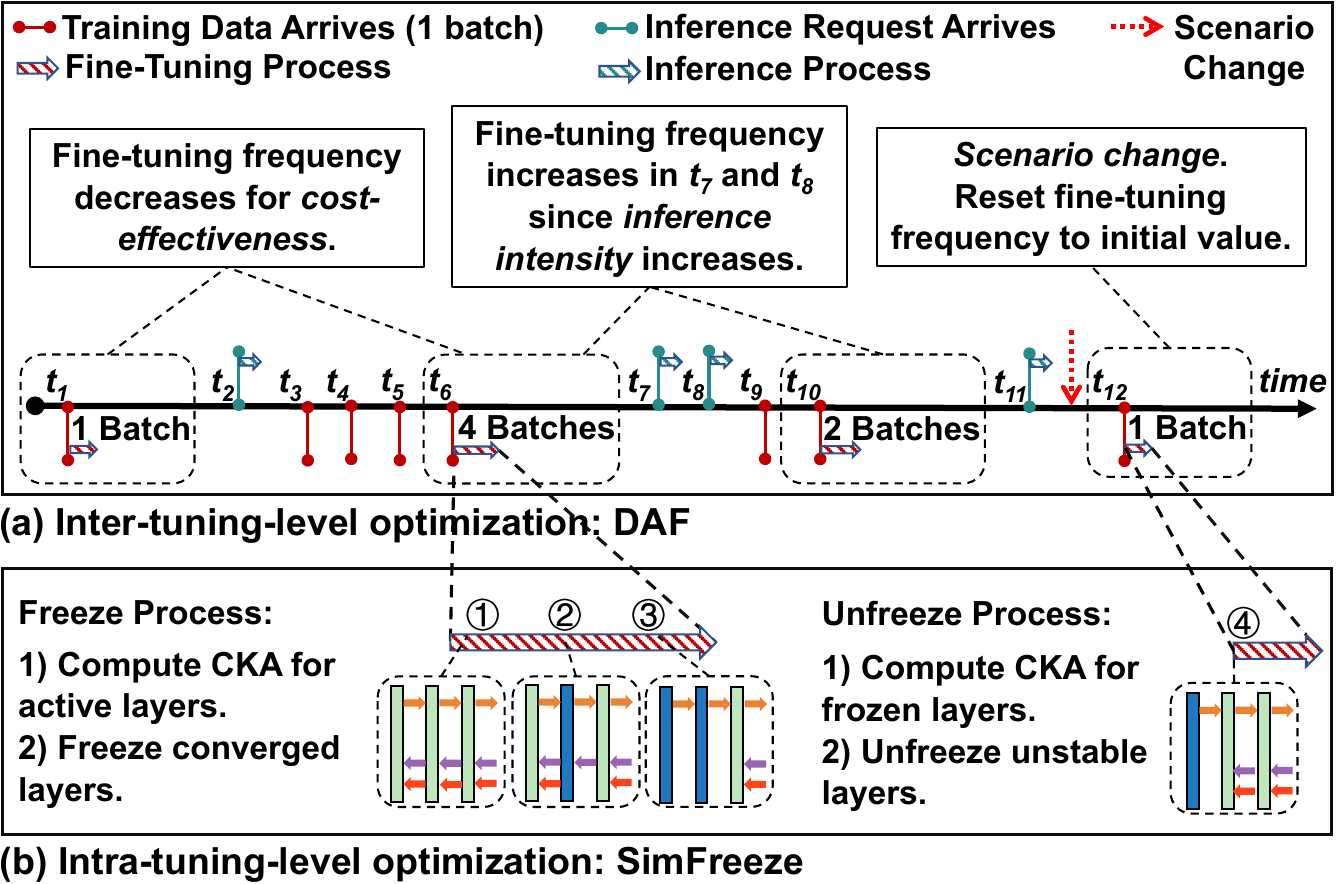}
\vspace{-0.22in}
\caption{Overview of EdgeOL. \ding{172}, \ding{173}, and \ding{174} in Figure \ref{fig:EdgeOL}b indicate the occurrence of freezing, matching the case 1, 2, and 3 in Figure \ref{fig:computation_freezing}, respectively. \ding{175} indicates the occurrence of unfreezing right after a scenario change.}
\label{fig:EdgeOL}
\end{figure}

Based on the insights, we propose {\tt EdgeOL}, an efficient online learning framework for edge devices.
Figure \ref{fig:EdgeOL} shows the overview of the {\tt EdgeOL} framework to achieve energy efficiency and high inference accuracy through i) inter-tuning and ii) intra-tuning optimizations. 
Specifically, for \textbf{inter-tuning}, we propose a novel \textbf{Dynamic and Adaptive Fine-tuning Frequency (DAF)} design that dynamically and adaptively adjusts the fine-tuning frequency to reduce the execution time and energy consumption (Section \ref{sec:scheme_freq}).
For \textbf{intra-tuning}, we propose a \textbf{similarity-guided freezing (SimFreeze)} method to automatically freeze/unfreeze layers during online learning to save computation costs while improving accuracy (Section \ref{sec:scheme_freezing}).
Moreover, {\tt EdgeOL} can also use unlabeled data through semi-supervised learning techniques to enhance model performance without the need for extensive labeled data (Section {\ref{sec:scheme_semi}}).
The {\tt EdgeOL} optimization design is described in Algorithm~\ref{algo} with terminology and abbreviations listed in Table~\ref{abbr}.

\subsection{Dynamic and Adaptive Fine-tuning Frequency (DAF)}
\label{sec:scheme_freq}

\subsubsection{Key Design Factors}
\label{sec:scheme_freq_factor}
\hfill\break
\textbf{Cost-Effectiveness.} 
As discussed in Section~\ref{sec:moti_inter},
each time launching a fine-tuning round will inevitably introduce extra time and energy overheads.
Therefore, we must consider the cost-effectiveness. 
It indicates whether the potential model accuracy improvement of launching a fine-tuning round is worth the incurred overheads.




\textbf{Inference Intensity.}
Inference intensity is also closely related to the performance of online learning in real-world applications.
Since each incoming inference request is served using the model at that time with instantaneous model accuracy, it would be more desirable to keep the model up-to-date when the inference requests are frequent.
Therefore, we take the ``inference intensity" into account for adaptive fine-tuning frequency adjustment, improving the practicality of our method.

\textbf{Scenario Changes.} The model may undergo multiple deployment scenario changes during online learning.
It is crucial for the model to update quickly to adapt to these changes to deliver satisfactory results.

In general, the scenario changes can be identified by many different methods. For example, one can track the distribution difference of the data in streaming-in inference requests using methods such as Least-Squares Density Difference (LSDD)~\cite{gama2014survey}, Maximum Mean Discrepancy (MMD)~\cite{gretton2012kernel}, and many other out-of-distribution data detection approaches~\cite{wu2022energy, liu2020energy, hendrycks2018deep, chen2021atom, hsu2020generalized}. Or it can also be detected by a stand-alone sensor module in a comprehensive system (e.g., robotics system~\cite{lesort2020continual, zhao2021general}).
{\tt EdgeOL} is compatible with these detection methods, and it is not our focus in this work.

\subsubsection{DAF Design Principle}
\label{sec:scheme_freq_design}
\hfill \break
To take the above-mentioned design factors into account, the proposed DAF dynamically orchestrates the fine-tuning frequency based on training data availability, the trend of the model's validation accuracy, the intensity of inference requests, and the changes in the model deployment scenario.
It is important to emphasize that validation accuracy differs from inference accuracy of inference requests. Validation accuracy is obtained by evaluating the model on a validation dataset, which is formed by randomly separating a small portion ($\sim$5\%) of the streaming-in training data while maintaining the same data distribution~\cite{ponce2006dataset}.
We cannot use inference accuracy because, in real-world applications, the inference requests will not have the corresponding ground truth labels; thus, we use validation accuracy to indicate model performance.

Specifically, DAF controls the fine-tuning frequency by using a tunable variable $batches\_needed$. A fine-tuning round is triggered only if the available streaming-in training data reaches $batches\_needed$ (line 2 in Algorithm~\ref{algo}).
A larger $batches\_needed$ indicates a lower fine-tuning frequency, wh-
ere more fine-tuning rounds are delayed and merged.
In our design, the initial value of $batches\_needed$ is the same as immediate online learning (i.e., 1 batch). 
And we use the following principles to adaptively tune up/down the $batches\_needed$ during online learning.

\textbf{Tuning down the frequency considering cost-effec-
tiveness.}
In general, within one scenario, as the model gradually converges through multiple fine-tuning rounds during online learning, the cost-effectiveness of the fine-tuning rounds decreases (see Section~\ref{sec:accuracy-each-fine-tuning-round}). 
Therefore, the fine-tuning frequency should be gradually tuned down to remain cost-effective.
Specifically, since launching each fine-tuning round incurs similar overhead, we consider maintaining cost-effectiveness as letting each round achieve comparable accuracy improvements.
Accordingly, after a fine-tuning round, DAF estimates the amount of training data needed for the next round to achieve similar accuracy gains as the most recent round (lines 12 and 13 in Algorithm~\ref{algo}).

Inspired by prior works~\cite{bhardwaj2022ekya, peng2018optimus, mahajan2020themis}, this estimation is achieved by employing a logistic regression model~\cite{hosmer2013applied}, which extrapolate the accuracy improvement curve from data collected from prior fine-tuning rounds, as described by:
\begin{equation}
     A(t) = \frac{L}{1 + e^{-k(t - t_0)}} 
\end{equation}
where $A(t)$ is the estimated accuracy at training iteration $t$, $L$ represents the curve's asymptote representing the maximum achievable accuracy, $k$ is the average accuracy growth rate, $t_0$ is the inflection point where the accuracy improvement rate begins to decline, and $e$ is the base of the natural logarithm. 
DAF employs the Non-Negative Least Squares (NNLS) solver~\cite{scipy.optimize.nnls} to fit this accuracy curve to the (training iteration, validation accuracy) data points collected from previous fine-tuning rounds. Each data point encapsulates the number of training iterations the model has experienced and the achieved validation accuracy at that iteration.

This model is then used to predict future accuracy improvements.
This enables DAF to ensure sufficient accuracy gains in subsequent fine-tuning rounds, thereby maintaining cost-effectiveness. Typically, this is achieved by strategically increasing $batch\_needed$ (i.e., tuning down the fine-tuning frequency). Note that the accuracy might drop after a fine-tuning round due to the fluctuating nature of model training. In this case, DAF will use the accuracy gains in the previous round as the accuracy improvement target for the next round.

\textbf{Tuning up the frequency considering inference intensity.}
To effectively improve inference accuracy under an intensive inference period, we tune up the fine-tuning frequency in DAF via a popular logarithmic-based function~\cite{manaseer2009logarithmic}. 
It is calculated as $d=d*(1-1/log(d))$,
where $d$ represents the number of data batches needed to trigger a fine-tuning round.
If inference requests are intensive, the $batches\_needed$ will be quickly decreased, thereby tuning up the fine-tuning frequency (lines 16 to 19 in Algorithm \ref{algo}). 
We opt for a logarithmic-based function to decrease $batches\_needed$ because it provides a moderate adjustment compared to two other prevalent value-adjusting functions: exponential-based function~\cite{kwak2005performance} and additive-based function~\cite{dumas2002markovian}. The logarithmic-based function is less aggressive than the exponential-based function, yet more aggressive than the additive-based function.

\textbf{Resetting the frequency upon a scenario change.}
Once a scenario change is detected, to ensure a quick adaptation to the new scenario to deliver satisfactory results, we increase the model fine-tuning frequency by resetting the frequency to the initial value.

\begin{figure}[t]
\centering
    \includegraphics[width=0.93\columnwidth]{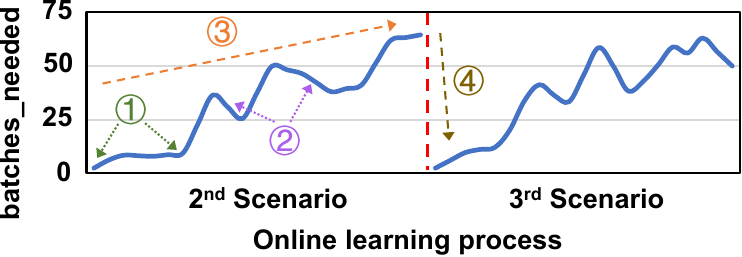}
\caption{An example of adaptive adjustment using DAF. The result is obtained by fine-tuning ResNet50 on the NC benchmark of the CORe50 dataset. The model is well-trained on the first scenario and then experiences subsequent scenarios. Here we show the 2nd and 3rd scenarios as an example. The red dotted line indicates a scenario change.}
\label{fig:freq}
\end{figure}

\subsubsection{Case study}
\hfill \break
Figure~\ref{fig:freq} shows a real example of how DAF adaptively adjusts the fine-tuning frequency in two consecutive scenarios.
From the figure, we have the following observations: \ding{172} shows that $batches\_needed$ remains a small value for several fine-tuning rounds. 
This is because significant accuracy improvements are achieved at the beginning of the online learning process in each scenario, and our DAF intends to keep a high fine-tuning frequency.
\ding{173} shows obvious decreases in $batches\_needed$ as the DAF responds to the intensive incoming inference requests at those moments. Moreover, DAF increases the fine-tuning frequency to keep the model up-to-date.
\ding{174} shows the overall increasing trend of $batches\_needed$ throughout the online learning process in each scenario since the model has generally converged, and DAF decreases the fine-tuning frequency (increases $batches\_needed$) to facilitate higher cost-effectiveness and reduce energy consumption.
\ding{175} shows a significant decrease in $batches\_needed$ upon a scenario change, as DAF increases the fine-tuning frequency by setting it to the initial value. This ensures quick model adaption in the new scenario.

\setlength{\textfloatsep}{1pt}
\begin{algorithm}[!t]
\caption{EdgeOL}
\label{algo}
\LinesNumbered
\footnotesize
    
    \textcolor{gray}{\# \emph{Fine-tuning}}
    
    \If{$batches\_ava \geq batches\_needed $}{
        \Call{trigger\_fine\_tuning}{$ $}\;
            
        \textcolor{gray}{\# \emph{SimFreeze}}
        
            \For {\rm every $freeze\_interval$ training iterations}{
                \For {\rm each active layer}{
                    \Call{cka\_calculation}{$ $}\;
                    
                    \If{$CKA\_variation \leq CKA\_TH$}{
                        \Call{freeze\_layer}{$ $}\;
                    }
                }
                $freeze\_interval \gets freeze\_interval \times (1-1/log(freeze\_interval))$ \;
                
            }
                
        
        \textcolor{gray}{\# \emph{Tuning down the fine-tuning frequency}}
        
        \If{fine-tuning ends}{
            $batch\_needed$ $\gets$ \Call{batch\_needed\_estimation}{$ $}\;
            
                
        }
     }
     
    \textcolor{gray}{\# \emph{Inference}}
    
    \textcolor{gray}{\# \emph{Tuning up the fine-tuning frequency}}
    
    \If{\textit{inference arrives}}{
        \Call{do\_inference}{$ $}\;
        \If{\textit{inference ends}}{
                $batches\_needed \gets batches\_needed \times (1-1/log(batches\_needed))$ \;
        }
    }
    
    \textcolor{gray}{\# \emph{Handling scenario changes}}
    
    \If{a scenario change is detected}{
            \Call{reset\_fine-tuning\_frequency}{$ $}\;
            \Call{update\_CKA\_test\_data}{$ $}\;
            \For {\rm each frozen layer}{
                \Call{comp\_cka\_with\_previous {\bf \&} new\_scenario\_data}{$ $}\;
                \If{$CKA\_variation \geq CKA\_TH$}{
                    \Call{unfreeze\_layer}{$ $}\;
                }
            }
    }
\end{algorithm}

\subsection{Similarity-Guided Freezing (SimFreeze)}
\label{sec:scheme_freezing}

We next design SimFreeze that adaptively freezes and unfreezes appropriate layers during online learning.
In general, as the model gradually converges as training proceeds within one scenario, SimFreeze identifies and freezes those converged layers. Upon encountering a scenario change, SimFreeze selectively resumes training on previously frozen layers that become unstable in the new scenario, facilitating a rapid and efficient adaptation to the changes.




\subsubsection{Utilizing self-representational similarity to guide layer freezing.}
\hfill\break
SimFreeze uses a layer's self-representational similarity to guide whether it can be frozen.
We consider the initial model before fine-tuning as the reference model.
We define the self-representational similarity of a layer as the degree of similarity between the output feature maps of a layer in the current model version and the output feature maps of that layer in the reference model.
As fine-tuning proceeds, the layers of the model are updated over time, and their self-representational similarity is also recorded.
When a layer's self-representational similarity is stabilized, then we consider the layer to have converged and it can be frozen.

To measure the self-representational similarity of two layers from two models, we use a widely-used metric \textit{Centered Kernel Alignment} (CKA)~\cite{kornblith2019similarity}.
The CKA value is obtained by comparing the output feature maps of two layers using the same input image batch.
It can be calculated as:
\begin{equation}
    CKA\left( X, Y \right) = \left\|Y^{\mathrm{T}} X\right\|_{\mathrm{F}}^{2} /\left(\left\|X^{\mathrm{T}} X\right\|_{\mathrm{F}}\left\|Y^{\mathrm{T}} Y\right\|_{\mathrm{F}}\right) \label{equa:CKA}
\end{equation}
where $X$ and $Y$ are the output feature maps from two layers, and $\left\|\cdot \right\|_{\mathrm{F}}^{2}$ represents the square of the Frobenius norm.
A higher CKA value represents that the two layers can generate more similar output feature maps using the same inputs.
Moreover, if the CKA value of a layer stabilizes as fine-tuning proceeds, we consider this layer to have converged.

Instead of comparing to the initial model, an intuitive alternative is to use models from prior training iterations as reference models. In this case, we can compare the current model to the model in earlier training iterations to monitor the CKA variation trend. However, this requires regular updates of the reference model as fine-tuning proceeds, leading to an increase in memory writes and consequently higher energy consumption. As such, we would not adopt this method in our framework.

Within a scenario, we collect the first arrived training data batch as the CKA test data for that scenario. 
The CKA test data will be used as the input for the models to generate output feature maps of each layer.
As shown in Algorithm \ref{algo} (lines 5 to 7), 
periodically (e.g., every 200 iterations),
we calculate the CKA and check the self-representational similarity for each active (non-frozen) layer, which is the first step of the freezing process in Figure \ref{fig:EdgeOL}b.
For each active layer, we compare the CKA value of the current model to the CKA value calculated last time.
The layers whose CKA variations are below the threshold (e.g., 1\%) are considered converged and will be frozen (lines 8 and 9 in Algorithm \ref{algo}), as the second freezing step in Figure \ref{fig:EdgeOL}b.

\begin{figure}[t]
\centering
    \includegraphics[width=0.95\columnwidth]{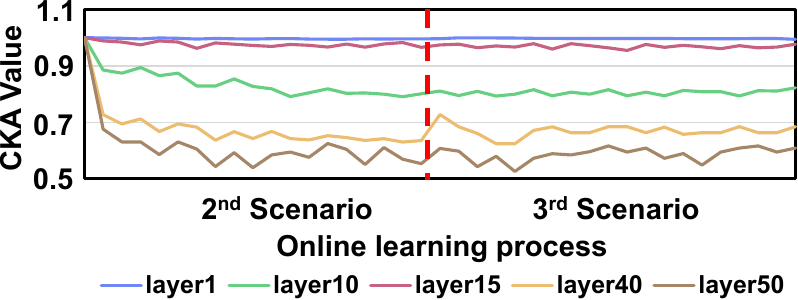}
\caption{CKA variation curve as fine-tuning proceeds. The result is obtained by fine-tuning ResNet50 on the NC benchmark. The model is well-trained on the first scenario and then experiences subsequent scenarios. Here we show the CKA value in the 2nd and 3rd scenarios as an example.}
\label{fig:CKA_curve_front}
\end{figure}

Figure \ref{fig:CKA_curve_front} shows an example of the trend of CKA values of layers 1, 10, 15, 40, and 50 as fine-tuning proceeds.
From the figure, we have the following observations.
First, different layers require a different number of training iterations to converge. For example, layer 1 converges at the very beginning, while layer 50 fluctuates all the time.
Moreover, it is interesting to observe that later layers can converge faster than earlier layers (e.g., layer 15 vs. layer 10). This is due to residual connections in the model network architecture, making some later layers behave like earlier layers~\cite{veit2016residual}.
These observations show the feasibility and necessity of freezing layers in an adaptive manner rather than sequentially from front to back.
Besides, we can also observe that once a layer has converged, its CKA value will remain stable and will not fluctuate significantly again within the same scenario.
Therefore, if a layer is frozen, then it would be good to keep it frozen for higher energy efficiency unless there are changes in the model deployment scenario. 


\subsubsection{Unfreezing layers upon scenario changes.}
\hfill\break
Once a scenario change is detected, we need to resume training on certain frozen layers to ensure a quick adaptation.
However, it may not be necessary to unfreeze all the frozen layers since some front layers are task-agnostic (discussed in Section{~\ref{sec:moti_intra}}).
To decide which layers to unfreeze, we first update the CKA test data with the new scenario data (line 23 in Algorithm~\ref{algo}).
Then, for each frozen layer, we compute its CKA using both original and new scenario CKA test data (lines 24 to 25 in Algorithm~\ref{algo}), as the first step of the unfreezing process in Figure \ref{fig:EdgeOL}b.
If the CKA variation of a layer exceeds the threshold after the scenario change, it indicates that the feature extraction ability of that layer for new scenario data and previous scenario data is significantly different.
In this case, we unfreeze that unstable layer to allow it to adapt to the new scenario (lines 26 to 27 in Algorithm~\ref{algo}), as the second unfreezing step in Figure \ref{fig:EdgeOL}b.

In the example in Figure \ref{fig:CKA_curve_front}, layers 1, 10, 15, and 40 are stable and can be frozen before the scenario change. After the scenario change, layers 1, 10, and 15 remain stable, while layer 40 becomes unstable. Consequently, we should resume training on layer 40 to let it quickly adapt to the new scenario.



\subsubsection{Dynamic interval to conduct freezing processes.}
\hfill\break
SimFreeze continuously conducts freezing processes (i.e., tracks the CKA variation and freezes the converged layers). 
However, conducting the freezing process too frequently (e.g., every training iteration) can lead to excessive overhead, as CKA calculation requires time and energy.
Therefore, we opt for periodically tracking the CKA variation of active layers and freezing the converging ones. Further, we propose using a logarithmic-based function that progressively decreases the interval before subsequent freezing process after each freezing process (line 10 in Algorithm \ref{algo}).
This approach takes into consideration that more and more layers converge as fine-tuning proceeds. For example, layer 10 in Figure \ref{fig:CKA_curve_front} has not converged at the early training stage but gradually converged.
The log-based function is calculated as $n=n*(1-1/log(n))$, where $n$ represents the interval to conduct a freezing process.

\subsection{Utilizing Unlabeled Data}\label{sec:scheme_semi}

It is possible that a portion of the streaming-in training data arrives without labels, posing a challenge for traditional supervised fine-tuning processes.
To address this issue and fine-tune the model in this case, {\tt EdgeOL} adopts a semi-supervised learning technique~\cite{chapelle2009semi} to make use of both labeled and unlabeled data.
In each fine-tuning round, {\tt EdgeOL} first fine-tunes the model using unlabeled data via the self-supervised learning~\cite{chen2020simple} to improve the feature extraction ability of the model.
Then, {\tt EdgeOL} fine-tunes the model by supervised learning using the labeled data to improve the model's performance in the target task (e.g., image classification for a particular dataset). 
This approach ensures {\tt EdgeOL}'s robust performance and adaptability.

\begin{figure*}[ht]
  \centering
\begin{minipage}{0.49\textwidth}
    \includegraphics[width=1\columnwidth]{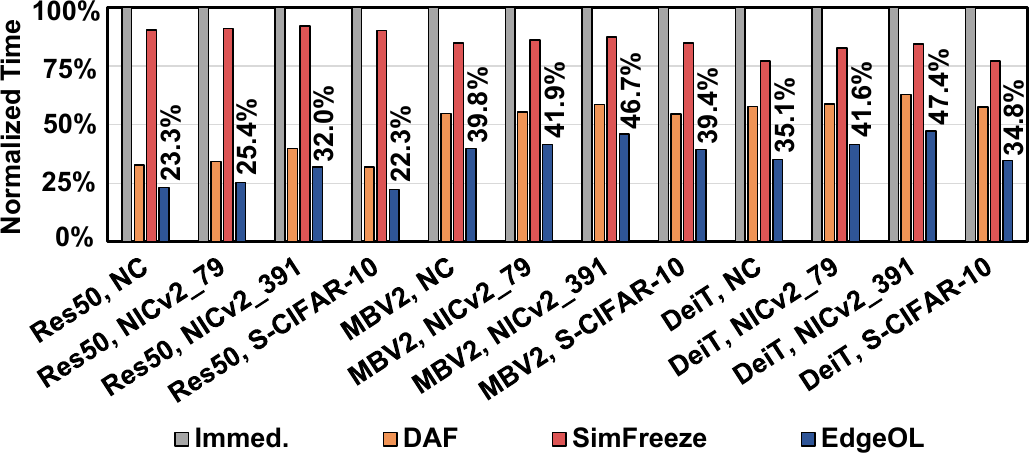} 
    \caption{Overall fine-tuning execution time.}
    \label{fig:eval_overall_time}
\end{minipage}
\hspace{4.5pt}
\begin{minipage}{0.49\textwidth}
    \includegraphics[width=1\columnwidth]{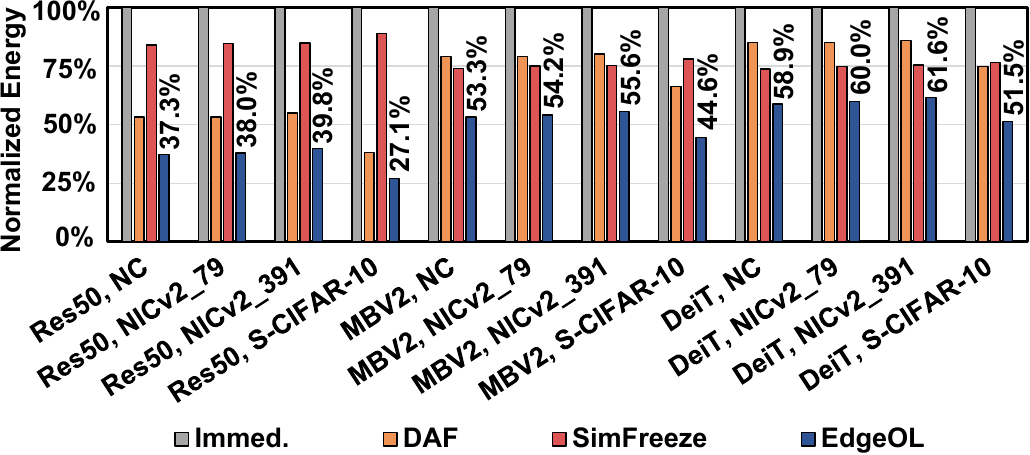}
    \caption{Overall fine-tuning energy consumption.}
    \label{fig:eval_overall_energy}
\end{minipage}
\vspace{0.06in}
\end{figure*}

\section{Evaluation}
\label{sec:eval}



In this section, we evaluate the proposed {\tt EdgeOL} framework using popular online learning workloads from both CV and NLP domains.

\subsection{Experimental Setup} 
\label{sec:motivation_inter_setup}
\textbf{Platform:}
We use the NVIDIA Jetson Xavier NX as our platform and choose the \textit{15W 6-Core} power mode with maximum GPU speed.

\textbf{Model and dataset:}
In the CV domain, we employ two popular CNN models ResNet50 (Res50 for short) and MobileNetV2 (MBV2 for short), as well as a vision transformer model DeiT (tiny version)~\cite{pmlr-v139-touvron21a}.
We employ three benchmarks NC, NICv2\_79, and NICv2\_391~\cite{lomonaco2020rehearsal} from the CORe50 dataset for evaluation, which contain 9, 79, and 391 scenarios, respectively.
CORe50 is a popular dataset that is widely used in several prior continuous online learning works~\cite{logacjov2021learning, mandivarapu2020self, graffieti2022continual, pellegrini2021continual}.
In the NC benchmark, each scenario introduces new classes of data on top of existing classes.
On the other hand, in NIC benchmarks, each scenario can introduce either i) new classes of data, ii) instances of existing classes but with new patterns (e.g., different environmental conditions such as changes in illumination and background), or iii) a combination of both.
We also use another widely-used benchmark S-CIFAR-10~\cite{buzzega2020dark, wang2022sparcl} to evaluate {\tt EdgeOL}, where the CIFAR-10~\cite{krizhevsky2009learning} dataset is split into 5 scenarios, each consisting of 2 distinct data classes.
In the NLP domain, we employ the BERT-base model~\cite{kenton2019bert} and the 20News benchmark used in several prior works~\cite{kirkpatrick2017overcoming, ke2021adapting, ke2021achieving, sun2019lamol}, where the 20News dataset~\cite{lang1995newsweeder} is split into 10 scenarios, each containing 2 data classes.


\textbf{Fine-tuning Setting:}
In our experiments, the model is originally well-trained in the first scenario. In the subsequent scenarios that appear sequentially, it will be online fine-tuned with corresponding training data and meanwhile serve inference requests. The CopyWeights with Re-init (CWR) technique proposed by the CORe50 benchmark paper is by default applied in the experiments to mitigate the catastrophic forgetting problem~\cite{zenke2017continual, li2019learn}.

In each scenario during the entire online learning process, both the training data and inference requests arrive continuously over time.
The arrival granularity of training data is 1 batch each time and the training batch size is fixed to 16 to avoid out-of-memory errors. We assume a total of 500 inference requests across all scenarios.
The arrival rate for both the training data and inference requests follows a Poisson distribution to mimic real application scenarios~\cite{mattson2020mlperf}. 
We also provide a sensitivity study on different numbers of inference requests and different arrival distributions in Section~\ref{sec:eval_sen_distribution}. 
Each dataset contains training and testing sets, and a portion of training data (5\%) is randomly separated to form a validation dataset as discussed in Section~\ref{sec:scheme_freq_design}.

\textbf{Baseline and SOTA Comparisons:} We use immediate online learning as our baseline, where the models are fine-tuned once training data is available. It provides the highest accuracy over all the online learning strategies with fixed fine-tuning frequency (see Figure \ref{fig:moti_inter_acc} for details). 
We also compare {\tt EdgeOL} with state-of-the-art efficient training methods (Section \ref{sec:eval_comparison}), including layer freezing methods i) Egeria~\cite{wang2023egeria} and ii) SlimFit~\cite{ardakani2023slimfit}, iii) sparse training framework RigL~\cite{evci2020rigging}, and iv) efficient online learning framework Ekya~\cite{bhardwaj2022ekya}.

\textbf{Metrics:}
We use three metrics for evaluation: overall fine-tuning execution time, overall energy consumption, and average inference accuracy. 
The overall fine-tuning execution time and energy consumption refer to the total time and energy costs of all scenarios during the entire online learning process. 
They sum up the fine-tuning execution time and energy consumption of all fine-tuning rounds.
The average inference accuracy is the average of accuracies over all inference requests in all scenarios.
All reported results are the average over 5 runs using different random seeds. Unless otherwise stated, the accuracy results refer to the average inference accuracy.

\begin{table}[t]
	\centering
\footnotesize
\caption{Average inference accuracy of all methods.}
    \setlength{\tabcolsep}{4.8pt}
    \begin{tabular}{m{0.9cm} m{1.15cm} m{0.8cm}<{\centering} m{1.02cm}<{\centering} m{1.1cm}<{\centering} m{1.38cm}<{\centering}}
    \toprule
        \multirow{3}{*}{Model} & \multirow{3}{*}{Method} & \multicolumn{4}{c}{Benchmark} \\
        \cline{3-6}
        & & NC & NICv2\_79 & NICv2\_391 & S-CIFAR-10\\ 
        \midrule
            \multirow{4}{*}{Res50}  & Immed.	                     &   71.34                        &   66.85                          & 58.76                        & 86.41\\
                                    \cline{2-6}
                                    & DAF                            &	 71.17                        &   66.59                          & 58.47                        & 86.30\\
                                    & SimFreeze                      &   73.91                        &   69.23                          & 60.59                        & 88.24\\
                                    & EdgeOL                         &   73.73                        &   69.04                          & 60.43                        & 88.12\\
            \hline
            \multirow{4}{*}{MBV2}   & Immed.	                     &   68.46                        &   62.89                          & 50.65                        & 83.56\\
                                    \cline{2-6}
                                    & DAF                            &   68.11                        &   62.54                          & 50.49                        & 83.04\\
                                    & SimFreeze                      &   70.72                        &   65.15                          & 52.62                        & 85.34\\
                                    & EdgeOL                         &   70.31                        &   64.96                          & 52.41                        & 85.09\\
            \hline
            \multirow{4}{*}{DeiT}   & Immed.	                     &   69.12                        &   61.22                          & 51.62                        & 84.43\\
                                    \cline{2-6}
                                    & DAF                            &   68.95                        &   61.12                          & 51.44                        & 84.38\\
                                    & SimFreeze                      &   70.99                        &   63.11                          & 53.02                        & 85.91\\
                                    & EdgeOL                         &   70.69                        &   62.95                          & 52.77                        & 85.80\\
        \bottomrule
    \end{tabular}
\vspace{0.16in}
\label{table_acc_overall}	
\end{table}

\subsection{Main Results}
\label{sec:eval_overall}

\subsubsection{CV Tasks}
\label{sec:eval_overall_cv}
\hfill\break
Figures~\ref{fig:eval_overall_time},~\ref{fig:eval_overall_energy}, and Table~\ref{table_acc_overall} show the overall execution time, energy consumption, and average inference accuracy of the immediate online learning and our proposed frameworks in CV domain.
The execution time and energy consumption are normalized to $Immed.$

\begin{figure*}[ht]
  \centering
\begin{minipage}{.291\textwidth}
        \centering
    \footnotesize
 \captionof{table}{Computation cost of entire online learning process of the NC benchmark.}
    \begin{tabular}{m{1.2cm}<{\centering} m{1.35cm}<{\centering} m{1.35cm}<{\centering}}
	\toprule
	    \multirow{2}{*}{Method} & \multicolumn{2}{c}{Computation (TFLOPs)}\\
	    \cline{2-3}
		  & Res50 & MBV2 \\ 
		\midrule
            Immed.      	    & 4,746  & 367 \\
            \rowcolor[gray]{.9}EdgeOL           & 3,037   & 124 \\
		\bottomrule
	\end{tabular}
    \label{table_comp}
\end{minipage}
\hspace{4.5pt}
\begin{minipage}{.33\textwidth}
    \includegraphics[width=0.97\columnwidth]{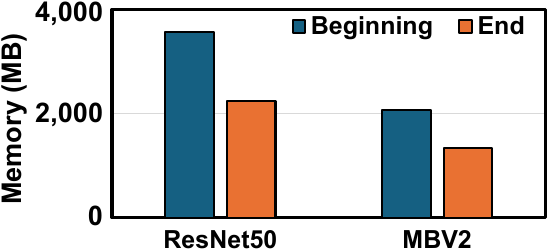} 
        \captionof{figure}{Memory usage at the beginning and the end of online learning. 
        }
        \label{fig:eval_comp_memory}
\end{minipage}
\hspace{4.5pt}
\begin{minipage}{.335\textwidth}
    \includegraphics[width=0.935\columnwidth]{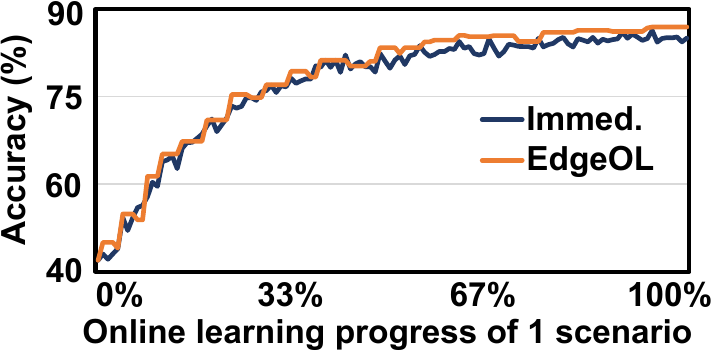}
        \captionof{figure}{Convergence of ResNet50 in one of the scenarios in NC benchmark.}
        \label{fig:eval_acc_curve}
\end{minipage}
\end{figure*}

\textbf{DAF.} 
As shown in Figures \ref{fig:eval_overall_time} and \ref{fig:eval_overall_energy}, DAF saves average 50\%, peak 68\% execution time, and average 31\%, peak 62\% energy compared to $Immed$.
These savings come from merging and delaying certain fine-tuning rounds, which can effectively reduce the execution overheads (by 92\% on average), including model loading, saving, and system initialization (e.g., model compilation).
As shown in Table~\ref{table_acc_overall}, despite the impressive gains in time and energy efficiency, DAF only incurs a minor 0.22\% accuracy drop compared to $Immed$.
This is because it adaptively determines the fine-tuning frequency that fits the current situation.

\textbf{SimFreeze.}
SimFreeze reduces average 15\%, peak 23\% execution time and saves average 22\%, peak 26\% energy compared to $Immed.$, as shown in Figures~\ref{fig:eval_overall_time} and \ref{fig:eval_overall_energy}.
These gains stem from the 35\% average savings in model computation (i.e., forward and backward propagation) through layer freezing.
Notably, SimFreeze also delivers significantly higher accuracy, a 1.96\% average increase over $Immed.$, as shown in Table~\ref{table_acc_overall}.
The reasons are two-fold:
First, SimFreeze accelerates model convergence (shown in Figure \ref{fig:eval_acc_curve}) as freezing layers reduce the number of model weights being trained. 
Second, SimFreeze avoids excessive adaptation to training data by freezing well-trained layers.

\textbf{EdgeOL.}
{\tt EdgeOL} combines DAF and SimFreeze.
From Figures \ref{fig:eval_overall_time}, \ref{fig:eval_overall_energy}, and Table~\ref{table_acc_overall}, compared to $Immed.$,
{\tt EdgeOL} saves average 64\%, peak 78\% execution time and average 52\%, peak 73\% energy, and improves accuracy by an average of 1.75\%.
Note that {\tt EdgeOL} shows more time and energy savings in NC and S-CIFAR-10 benchmarks, as their scenario changes are less frequent (8 and 4 vs. 78 and 390), allowing greater optimization potential in both inter- and intra-tuning.
\textbf{Computation Cost and Memory Usage.}
Table \ref{table_comp} shows the computation cost reduction.
Note that computation cost reduction comes from SimFreeze, as DAF only delays and merges fine-tuning rounds.
{\tt EdgeOL} also saves memory since freezing layers can reduce the intermediate data generated during the computation. 
As shown in Figure \ref{fig:eval_comp_memory},
{\tt EdgeOL} can reduce the memory usage by 40\% for ResNet50 and MobileNetV2.

\textbf{Model Convergence Speed.}
Figure \ref{fig:eval_acc_curve} plots the model convergence during online learning in one of the scenarios. We observe that our {\tt EdgeOL} helps the model converge faster as layer freezing effectively reduce the number of model weights being trained, leading to a higher accuracy compared to immediate online learning.

{\bf Overheads.}
The major overhead of {\tt EdgeOL} is the CKA calculation in SimFreeze. 
This overhead is introduced by i) a forward propagation using a batch of data to get the output feature maps, ii) CKA calculation for active layers using the obtained output feature maps.
Fortunately, many layers will be frozen as training proceeds, so the computation of CKA decreases over time. Specifically, in our evaluation, SimFreeze incurs $<$2\% additional energy for CKA computation, a minor amount when compared to 52\% energy benefit from {\tt EdgeOL}. All the reported results have included all the overhead.

\begin{table}[htbp]
    \footnotesize
    \setlength\tabcolsep{5pt}
	\centering
\caption{Experimental results in NLP workloads.}
	\begin{tabular}{m{1.5cm}<{\centering} m{1.6cm}<{\centering} m{1.7cm}<{\centering} m{1.7cm}<{\centering}}
	\toprule
		 Method & Accuracy (\%) & Time (minute) & Energy (Wh) \\
		
		\hline
            Immed.                              & 65.43   & 329      & 64.58            \\
            \cline{1-4}
            DAF	                                & 65.11   & 193      & 50.97             \\
            SimFreeze                           & 67.27   & 248      & 43.70             \\
            EdgeOL                              & 66.95   & 110      & 30.02             \\
        \bottomrule
	\end{tabular}

\label{table_NLP}
\end{table}

\subsubsection{NLP Tasks} 
\label{sec:eval_overall_nlp}
\hfill\break
We further evaluate the {\tt EdgeOL} framework on NLP tasks to showcase its generalizability.
As shown in Table \ref{table_NLP}, when compared to the immediate online learning approach, {\tt EdgeOL} offers a reduction of 67\% in execution time and 54\% in energy consumption, while increasing the accuracy by 1.52\%.
These results demonstrate the generalizability and superiority of {\tt EdgeOL}.

\begin{table}[htbp]
\setlength\tabcolsep{1.5pt}
	\centering
\footnotesize
\caption{Comparison with SOTA efficient learning methods.}
    \begin{tabular}{m{0.9cm}<{\centering} m{1.6cm}<{\centering} m{1.22cm}<{\centering} m{1.45cm}<{\centering} m{1.22cm}<{\centering} m{1.45cm}<{\centering}}
	\toprule
	    \multirow{2}{*}{Model} & \multirow{2}{*}{Method} & \multicolumn{2}{c}{NC} & \multicolumn{2}{c}{NICv2\_391}\\
	    \cline{3-4} \cline{5-6}
		 & & Acc. (\%) & Energy (Wh) & Acc. (\%) & Energy (Wh) \\
		
		\hline
            \multirow{6}{*}{Res50}  & DAF (base)                             & 71.17    & 61.54     & 58.47      & 78.36 \\
                                    & Egeria~\cite{wang2023egeria}           & 71.41    & 52.05     & 57.18      & 68.61   \\
                                    & SlimFit~\cite{ardakani2023slimfit}     & 72.26	& 53.46	    & 58.41	     & 69.32   \\
                                    & RigL~\cite{evci2020rigging}            & 70.97	& 51.76	    & 57.93	     & 70.16 \\
                                    & Ekya~\cite{bhardwaj2022ekya}           & 73.57	& 55.45	    & 57.58	     & 68.02  \\
                                    &\cellcolor[gray]{.9}EdgeOL	             & \cellcolor[gray]{.9}73.73    & \cellcolor[gray]{.9}43.15     & \cellcolor[gray]{.9}60.43      & \cellcolor[gray]{.9}56.79 \\

        \hline
            \multirow{6}{*}{MBV2}   & DAF (base)                             & 68.46    & 19.18     & 50.49     & 24.37 \\
                                    & Egeria~\cite{wang2023egeria}           & 69.49	& 15.91	    & 50.63	    & 20.69 \\
                                    & SlimFit~\cite{ardakani2023slimfit}     & 67.88	& 15.85	    & 49.69	    & 21.35   \\
                                    & RigL~\cite{evci2020rigging}            & 68.45	& 17.57	    & 50.12	    & 21.49 \\
                                    & Ekya~\cite{bhardwaj2022ekya}           & 68.34	& 14.80	    & 52.54	    & 19.33  \\
                                    &\cellcolor[gray]{.9}EdgeOL  	         & \cellcolor[gray]{.9}70.31    & \cellcolor[gray]{.9}12.90      & \cellcolor[gray]{.9}52.41      & \cellcolor[gray]{.9}16.91 \\

        \hline
            \multirow{5}{*}{DeiT}   & DAF (base)                             & 68.95    & 71.86    & 51.44      & 91.89 \\
                                    & Egeria~\cite{wang2023egeria}           & 69.41	& 63.02	   & 51.56	    & 78.26   \\
                                    & SlimFit~\cite{ardakani2023slimfit}     & 68.79	& 61.01	   & 50.93	    & 77.80   \\
                                    & RigL~\cite{evci2020rigging}            & 68.48	& 65.25	   & 51.08	    & 81.80 \\
                                    & Ekya~\cite{bhardwaj2022ekya}           & 68.96	& 61.04	   & 51.06	    & 75.51 \\
                                    &\cellcolor[gray]{.9}EdgeOL  	         & \cellcolor[gray]{.9}70.69    & \cellcolor[gray]{.9}49.66     & \cellcolor[gray]{.9}52.77      & \cellcolor[gray]{.9}65.74 \\
    
    \bottomrule
	\end{tabular}
\label{compare_freezing}
\end{table}


\subsection{Comparison with State-of-the-art Efficient Learning Methods}
\label{sec:eval_comparison}
We compare {\tt EdgeOL} with state-of-the-art efficient training methods, including layer freezing methods i) Egeria~\cite{wang2023egeria} and ii) SlimFit~\cite{liu2021autofreeze}, iii) sparse training framework RigL~\cite{evci2020rigging}, and iv) efficient online learning framework Ekya~\cite{bhardwaj2022ekya}.
Results are presented in Table \ref{compare_freezing}.
Since all these methods do not consider optimizations of inter-tuning, which significantly limits their benefits in efficiency and accuracy.
For a thorough comparison, we integrate our inter-tuning optimization, DAF, into all methods with identical configurations.
Table \ref{compare_freezing} shows that even with DAF integration, {\tt EdgeOL} still consistently outperforms all these methods, providing 2.1$\times$, 2.2$\times$, 2.8$\times$, and 2.0$\times$ energy savings, respectively, while delivering 1.78\%, 2.18\%, 2.33\%, and 1.50\% higher accuracy.

{\tt EdgeOL} outperforms Egeria due to its more flexible and finer-grained layer-freezing approach.
Specifically, {\tt EdgeOL} assesses layers individually rather than in modules (i.e., layer blocks), and it freezes all identified converged layers without forcing layers to be frozen sequentially from front to back. Hence, it avoids overtraining already converged layers in the middle of a non-converged module or after a non-converged layer. 
Against SlimFit, {\tt EdgeOL}'s advantage lies in the use of a more reliable metric: layer representational similarity. This directly analyzes layer outputs, offering a more accurate assessment than indirect methods like monitoring weight update magnitudes, which SlimFit employs.
In contrast to RigL, {\tt EdgeOL} effectively addresses sparse training challenges, such as GPU underutilization and workload imbalance. Compared to Ekya, {\tt EdgeOL} eliminates the inefficiency of Ekya's trial-and-error method in training configuration (e.g., which layers to freeze), ensuring more effective and efficient performance improvements.

\begin{figure}[ht]
    \centering
    \includegraphics[width=0.9\columnwidth]{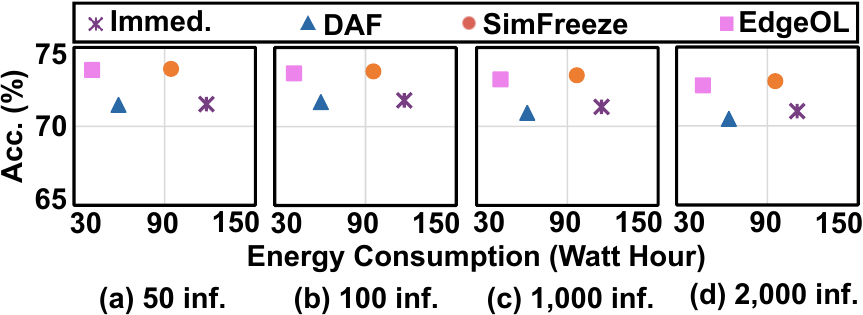}
    \vspace{-0.08in}
    \caption{Results under different number of inference requests.}
    \label{fig:eval_sen_infer}
    \vspace{-0.11in}
\end{figure}

\begin{figure}[ht]
    \centering
    \includegraphics[width=0.775\columnwidth]{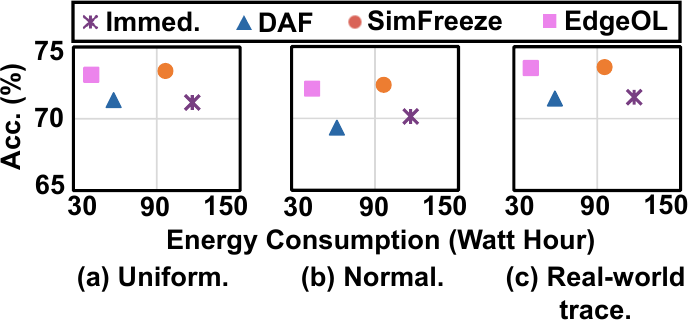} 
    \vspace{-0.09in}
    \caption{Results under different arrival distributions.}
    \label{fig:eval_sen_distribution}
    \vspace{-0.11in}
\end{figure}

\begin{figure}[ht]
    \centering
    \includegraphics[width=0.9\columnwidth]{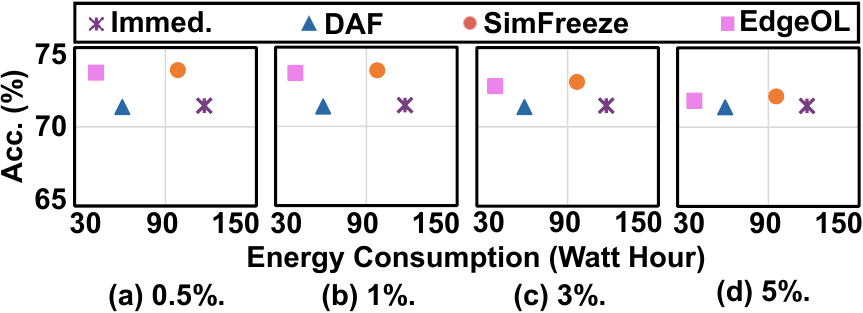}
    \vspace{-0.08in}
    \caption{Results under different CKA variation threshold.}
    \label{fig:eval_sen_cka}
\end{figure}

\subsection{Sensitivity Analysis}
\label{sec:eval_sen}
{\bf Number of inference requests.}
\label{sec:eval_sen_infer}
Figure \ref{fig:eval_sen_infer} shows the average inference accuracy and energy consumption under different numbers of inference requests.
Note that, all the inference requests are arriving following a Poisson distribution~\cite{mattson2020mlperf}.
All results in this section are obtained on ResNet50 and NC benchmark.
{\tt EdgeOL} consistently achieves higher accuracy than $Immed.$, while consuming significantly less energy. 
The figure also reveals that the energy saving offered by {\tt EdgeOL} increases as the total number of inference requests decreases.
This occurs because when the number of inference requests decreases, {\tt EdgeOL} (achieved by DAF) will decrease the fine-tuning frequency, thereby reducing the energy from execution overheads such as system initialization, as explained in Section \ref{sec:moti_inter}.

{\bf Data Arrival distribution.}
\label{sec:eval_sen_distribution}
In addition to the Poisson distribution, we also evaluate {\tt EdgeOL} under different arrival distributions for both training data and inference requests, including the uniform distribution~\cite{kuipers2012uniform}, normal distribution~\cite{altman1995statistics}, and a real-world trace from Video Timeline Tags dataset~\cite{huang2020multimodal}.
As depicted in Figure \ref{fig:eval_sen_distribution}, 
{\tt EdgeOL} consistently excels in both accuracy and energy consumption compared to $Immed$., showing that {\tt EdgeOL} is adept at handling a variety of situations with different data arrival distributions.

{\bf CKA variation threshold.}
In our experiments, a layer whose CKA variation is less than 1\% is considered converged (as mentioned in Section \ref{sec:scheme_freezing}). 
Additionally, we further evaluate the performance of {\tt EdgeOL} under various CKA variation thresholds. 
Figure \ref{fig:eval_sen_cka} shows that decreasing the threshold will lead to higher energy consumption and also higher accuracy. 
However, the accuracy saturates when the threshold is low enough (e.g., 1\%).

\begin{table}[htbp]
\setlength\tabcolsep{4.5pt}
	\centering
\footnotesize
\caption{Experimental results (NC benchmark) in semi-supervised learning.}
    \begin{tabular}{m{1.9cm}<{\centering} m{1.6cm}<{\centering} m{1.75cm}<{\centering} m{1.75cm}<{\centering}}
	\toprule
	    Model & Method & Accuracy (\%)  & Energy (Wh)\\
		
		\hline
            \multirow{2}{*}{ResNet50}   & Immed.                                &  60.28                        & 157.74    \\
                                        &\cellcolor[gray]{.9}EdgeOL	            &  \cellcolor[gray]{.9}61.64    & \cellcolor[gray]{.9}74.52  \\
        \hline
            \multirow{2}{*}{MobileNetV2}& Immed.                                &  55.33                        & 37.95 \\
                                        &\cellcolor[gray]{.9}EdgeOL	            &  \cellcolor[gray]{.9}56.87    & \cellcolor[gray]{.9}22.83  \\
        \hline    
            \multirow{2}{*}{DeiT}       & Immed.                                &  58.41                        & 138.05 \\
                                        &\cellcolor[gray]{.9}EdgeOL	            &  \cellcolor[gray]{.9}59.79    & \cellcolor[gray]{.9}87.79  \\
    
    \bottomrule
	\end{tabular}
\label{table_semi}
\end{table}

\subsection{Semi-supervised Learning}
Next, we evaluate the ability of our {\tt EdgeOL} to utilize unlabeled data by applying semi-supervised learning.
We choose the common configuration that only 10\% of the training data is labeled~\cite{zhai2019s4l, wang2022np}. 
As shown in Table \ref{table_semi}, compared to $Immed.$, {\tt EdgeOL} delivers 1.36\% higher accuracy and saves 43\% energy on average.
These results demonstrate that {\tt EdgeOL} works well in semi-supervised learning cases. 
This is because both DAF and SimFreeze are robust to the insufficient labeled data as i) DAF only needs a very small amount of labeled validation data to get the validation accuracy to adjust the fine-tuning frequency and ii) SimFreeze freezes layers by self-representational similarity, which can be acquired without data labels.


\begin{table}[h]
\setlength\tabcolsep{1pt}
	\centering
\footnotesize
\caption{Average inference accuracy when quantization is applied. The results are obtained on ResNet50.}
    \begin{tabular}{m{1.9cm}<{\centering} m{1.5cm}<{\centering} m{1.5cm}<{\centering} m{1.5cm}<{\centering} m{1.5cm}<{\centering}}
	\toprule
	    \multirow{2}{*}{Method} & \multicolumn{2}{c}{NC} & \multicolumn{2}{c}{NICv2\_79}\\
	    \cline{2-3} \cline{4-5}
		    &     8-bit & 32-bit & 8-bit & 32-bit\\ 
		
		\hline
            Immed.                           & 70.72   & 71.34   & 58.28   & 58.76   \\
            \rowcolor[gray]{.9}EdgeOL	     & 73.01   & 73.73   & 60.20   & 60.43   \\
    
    \bottomrule
	\end{tabular}
\label{table_quantization}
\end{table}

\subsection{Compatibility with Quantization}
\label{sec:eval_compatibility}
We also evaluate the compatibility of our {\tt EdgeOL} with quantization technique~\cite{stock2020training, gupta2015deep}.
We apply 8-bit fixed-point quantization to weights, activations, the gradient of weights, and the gradient of activations.
Following the prior works, we compare the accuracy results since the simulated quantization-aware training is used~\cite{zhou2016dorefa,wang2018training,zhu2020towards,zhao2021distribution}.
Table \ref{table_quantization} shows that {\tt EdgeOL} outperforms immediate online learning in 32-bit floating-point baselines with an accuracy improvement of 2.03\%. On the other hand, when employing 8-bit quantization, {\tt EdgeOL} achieves a 2.11\% higher accuracy.
These results suggest that {\tt EdgeOL}'s advantages are maintained when quantization techniques are used, demonstrating compatibility and robustness.

\section{Related Works and Discussion}
\label{sec:related}

A number of approaches have been proposed to reduce the computation costs of DNN models, thereby reducing energy and execution time.
E2Train~\cite{wang2019e2} proposes to drop mini-batches randomly, skip layers selectively, and use low-precision back-propagation during training to reduce the computation costs. \cite{DBLP:conf/iclr/YouL0FWCBWL20} designs a low-cost method to train the small but critical subnetworks to achieve the same accuracy as the original neural networks. 
\cite{hu2022lora} proposes to use lightweight low-rank matrices to adapt the weights of original models, slightly sacrificing model representational power to reduce the training costs.
However, these and most other prior works focus on offline learning.

For online learning, there are some works proposed to optimize particularly for online video analytics applications~\cite{khani2023recl, bhardwaj2022ekya}.
Specifically, RECL~\cite{khani2023recl} maintains a model zoo and uses the streaming-in training data to fine-tune these models, where the most appropriate model will be selected for inference in different scenarios.
Ekya~\cite{bhardwaj2022ekya} strategically schedules the resources among the training and inference workloads of co-running applications to achieve higher inference accuracy.
Due to the continuous and regular nature of video streaming in those applications, these works typically divide the online learning process into multiple short windows (e.g., 200 seconds) and conduct online fine-tuning in each window in a fixed-frequency manner.
Some other methods are proposed to filter important data for training to minimize the cost~\cite{panda2016conditional, wu2021enabling}, which can effectively reduce training costs.
Moreover, \cite{ma2023cost} presents a system runtime designed to dynamically configure the episodic memory hierarchy (HEM), where HEM is critical for improving the model performance during online learning. This runtime effectively optimizes both accuracy and energy efficiency.
Nonetheless, it is important to emphasize that our approach is complementary to these approaches since we focus on determining the moment to trigger fine-tuning adaptively and freezing layers selectively.
We will investigate the incorporation of the above methods in our future works.

\section{Conclusion}
\label{sec:conclusion}
In this paper, we design an efficient and accurate online framework for edge devices, namely {\tt EdgeOL}. 
It addresses requirements for both \textit{adaptiveness} and \textit{energy efficiency} for efficient online learning from both inter- and intra-tuning levels. 
Our experiments show that {\tt EdgeOL} significantly reduces training time and energy consumption while simultaneously improving inference accuracy.


\bibliographystyle{plain}
\bibliography{references}

\end{document}